\documentclass[10pt,twocolumn,letterpaper]{article}

\usepackage{iccv}
\usepackage{times}
\usepackage{epsfig}
\usepackage{graphicx}
\usepackage{amsmath}
\usepackage{amssymb}

\usepackage{subfigure}
\usepackage{stackrel}
\usepackage{wasysym}
\usepackage{dsfont}
\usepackage{enumitem}
\usepackage{bm}
\usepackage{booktabs}
\usepackage{multirow}
\setlist{nolistsep}
\usepackage[hang,flushmargin]{footmisc} 

\usepackage[dvipsnames]{xcolor}
\usepackage{tabu}
\usepackage{longtable}
\usepackage{color}
\usepackage{colortbl}

\definecolor{tabcolor1}{rgb}{0.556,0.847,0.973}
\definecolor{textcolor1}{rgb}{0.60,0.73,0.34}
\definecolor{textcolor3}{rgb}{0.255,0.412,0.882}
\definecolor{textcolor2}{rgb}{0.725,0.459,0.710}
\definecolor{layercolor1}{rgb}{0.255,0.412,0.882}

\newtheorem{proposition}{Proposition}

\newcommand*\xbar[1]{%
  \hbox{%
    \vbox{%
      \hrule height 0.5pt 
      \kern0.5ex
      \hbox{%
        \kern-0.1em
        \ensuremath{#1}%
        \kern-0.1em
      }%
    }%
  }%
}

\usepackage[pagebackref=true,breaklinks=true,letterpaper=true,colorlinks,bookmarks=false]{hyperref}

\usepackage[hyphenbreaks]{breakurl}
\iccvfinalcopy 

\def\iccvPaperID{733} 

\ificcvfinal\fi
\begin{document}

\title{Is Second-order Information Helpful for  Large-scale  Visual Recognition?}

\author{Peihua Li$^{1}$, Jiangtao Xie$^{1}$, Qilong Wang$^{1}$, Wangmeng Zuo$^{2}$\\
$^{1}$Dalian University of Technology, $^{2}$Harbin Institute of Technology\\
{\tt\small peihuali@dlut.edu.cn,\{jiangtaoxie,qlwang\}@mail.dlut.edu.cn,wmzuo@hit.edu.cn}
}

\maketitle
\thispagestyle{empty}

\begin{abstract}
By stacking layers of convolution and nonlinearity, convolutional networks (ConvNets)  effectively learn from low-level to high-level features and discriminative representations. Since the end goal of large-scale recognition is to delineate  complex  boundaries of thousands of classes,  adequate exploration of  feature distributions is important for realizing full potentials of ConvNets. However, state-of-the-art works concentrate only on deeper or wider architecture design, while rarely exploring feature statistics higher than  first-order. We take a step towards addressing this problem. Our method consists in  covariance pooling,  instead of the most commonly used first-order pooling, of high-level convolutional features. The main challenges involved are robust covariance estimation given a small sample of large-dimensional features and usage of the manifold structure of covariance matrices. To address these challenges, we present a Matrix Power Normalized Covariance (MPN-COV) method. We develop forward and backward propagation formulas regarding the nonlinear matrix functions such that MPN-COV can be trained end-to-end. In addition, we analyze both qualitatively and quantitatively its advantage over the well-known Log-Euclidean metric. On the  ImageNet 2012  validation set, by combining  MPN-COV  we achieve over 4\%, 3\% and 2.5\% gains for AlexNet, VGG-M and VGG-16, respectively; integration of MPN-COV into 50-layer ResNet outperforms ResNet-101 and is comparable to ResNet-152. The source code will be available on the project page: \href{http://www.peihuali.org/MPN-COV}{http://www.peihuali.org/MPN-COV}.
\end{abstract}

\section{Introduction}\label{section:introduction}

Since  outperforming significantly the classical, shallow classification framework, deep convolutional  networks (ConvNets)~\cite{nips2012cnn} have triggered fast growing interests and achieved great advance in  large-scale visual recognition~\cite{Simonyan15,Szegedy_2015_CVPR,He_2016_CVPR}.\let\thefootnote\relax\footnote{The work was supported by National Natural Science Foundation of China (No. 61471082). Peihua Li is the corresponding author. } The ConvNet architecture~\cite{LeNet1989} renders learning of  features, representations and classification in an end-to-end manner, superior to the classical Bag of Words (BoW)~\cite{Lazebnik:2006:BBF} architecture  where these components are separately optimized, independent of each other. The large-scale, labeled ImageNet dataset~\cite{Russakovsky2015} and high computing capability of GPUs contribute to successful training of increasingly wider and deeper ConvNets.

The ConvNet model, which starts from the raw color images as inputs, learns progressively the low-level, middle-level and high-level features from bottom, intermediate to top convolutional (conv.) layers~\cite{Zeiler2014}, obtaining  discriminative representations connected to  fully-connected (FC) layers. The gradient backprogation algorithm enables the classifier to learn decision boundaries delineating  thousands of classes in the space of large-dimensional features generally with complex distributions. Hence, for realizing full potentials of ConvNets, it is  important to adequately consider feature distributions. However,  most ConvNets concentrate only on designing wider or deeper architectures, rarely exploring statistical information higher than first-order. In the traditional classification paradigm where  sufficient labeled data are not available, high-order  methods combined with  ConvNet models pretrained on ImageNet dataset have achieved impressive recognition  accuracies~\cite{Cimpoi_2015_CVPR,Wang_2016_CVPR}.  In the small-scale classification scenarios, researchers have studied end-to-end methods, including DeepO$^{2}$P~\cite{Ionescu_2015_ICCV} and B-CNN~\cite{lin2015bilinear}, for exploiting second-order statistics in deep ConvNets~\cite{Ionescu_2015_ICCV,lin2015bilinear}.  As such, one interesting problem arising naturally is whether statistics higher than first-order is helpful for large-scale visual recognition.

In this paper, we  take a step towards addressing this problem. Motivated by~\cite{Ionescu_2015_ICCV,lin2015bilinear}, we perform covariance pooling of the last convolutional features rather than the  commonly used first-order pooling,  producing covariance matrices as global image representations. The main challenges involved are robust covariance estimation provided only with a small sample of large-dimensional features and usage of the manifold structure of the covariance matrices. Existing methods can not well address the two problems, producing unsatisfactory improvement in the large-scale setting.  DeepO$^{2}$P adopts Log-Euclidean (Log-E) metric~\cite{LogMetricsSIAM06} for exploiting geometry of covariance spaces, which however brings side effect on covariance representations. B-CNN performs element-wise normalization, without considering the manifold of covariance matrices. For tackling the challenges, we propose a Matrix Power Normalized Covariance (MPN-COV) method. We show that MPN-COV amounts to robust covariance estimation; it also approximately exploits the geometry of covariance space while circumventing the downside of the well-known Log-E  metric~\cite{LogMetricsSIAM06}. As MPN-COV  involves  nonlinear matrix functions whose backpropagation is not straightforward, we develop the gradients associated with MPN-COV based on the matrix propagation methodology~\cite{IonescuVS15} for end-to-end learning. 

Our main contributions are summarized as follows. Firstly, we are among the first who  attempt to exploit statistics higher than  first-order for improving the large-scale  classification. We propose matrix power normalized covariance method for  more discriminative representations, and develop the forward and backward propagation formulas for the nonlinear matrix  functions, achieving end-to-end MPN-COV networks. Secondly, we provide  interpretations of MPN-COV from statistical, geometric and  computational points of view, explaining the underlying mechanism that MPN-COV can address the aforementioned challenges. Thirdly, on the ImageNet 2012 dataset, we thoroughly evaluate  MPN-COV  for validating our mathematical derivation and understandings, obtaining competitive improvements over its first-order  counterparts under a variety of  ConvNet architectures.

\section{Related Work}\label{section:related-work}

The statistics higher than first-order has been successfully used in both classical and deep learning based classification scenarios. In the area of low-level patch descriptors, local Gaussian descriptors has demonstrated better performance than  descriptors  exploiting zeroth- or first-order statistics~\cite{LE2MG}. Fisher Vector (FV) makes use of the first- and second-order statistics, reporting  state-of-the-art results based on hand-crafted features~\cite{sanchez}. The locality-constrained affine subspace coding~\cite{Li-CVPR-2015} proposed to  use  Fisher Information matrix for improving classification performance.  By adopting features computed from pretrained ConvNets, FV considerably improves recognition accuracy over using  hand-crafted features  on small-scale datasets~\cite{Cimpoi_2015_CVPR}.  Wang et al.~\cite{Wang_2016_CVPR} present global Gaussian distributions as image representations for material recognition using the convolutional features from pretrained ConvNets.  In~\cite{Cimpoi_2015_CVPR,Wang_2016_CVPR},  feature design, image representation and classifier training are not jointly optimized. Different from them, we propose end-to-end deep learning to exploit the second-order statistics for improving large-scale visual recognition. 

In image classification the second-order pooling known as O$^{2}$P is  proposed in~\cite{Carreira:2012:SSS:2403272.2403306}. The O$^{2}$P   computes non-central, second-order moments which is subject to matrix logarithm for representing free-form regions. In the context of classical image classification, Koniusz et al.~\cite{Higher-order-PAMI2016} propose  second- and third-order  pooling of  hand-crafted features  or their coding vectors. For the goal of counteracting correlated burstiness due to non-i.i.d. data, they  apply  power normalization  of eigenvalues (ePN) to autocorrelation matrices or to the core tensors~\cite{Tensor-Decomposition} of the  autocorrelation tensors. In~\cite{WACV2017}, Higher-order Kernel (HoK) descriptor is proposed for action recognition in videos. HoK concerns pooling of higher-order tensors of probability scores from pretrained ConvNets in video frames, which are subject to ePN and then  fed to SVM classifiers. Our main differences from~\cite{Higher-order-PAMI2016,WACV2017} are (1) we  develop an end-to-end MPN-COV method in deep ConvNet architecture, and verify that statistics higher than  first-order is helpful for large-scale  recognition; (2) we provide statistical, geometric and computational  interpretations, explaining the mechanism underlying  matrix power normalization.

Ionescu et al.~\cite{Ionescu_2015_ICCV} present the theory of matrix backprogation which makes possible inclusion of structured, global layers into deep ConvNets. Furthermore, they propose DeepO$^{2}$P for  end-to-end, second-order pooling in deep ConvNets by Singular Value Decomposition (SVD).  B-CNN~\cite{lin2015bilinear}  aggregates the outer products of convolutional features from two networks. The resulting matrices undergo element-wise power normalization. Note that B-CNN produces second-order, non-central moments  when the two ConvNets involved share the same configuration.  Our MPN-COV is similar to DeepO$^{2}$P and B-CNN  but having clear distinctions. We show that matrix power normalization plays a key role for the second-order pooling to achieve competitive performance, instead of  matrix logarithm~\cite{LogMetricsSIAM06} or element-wise power normalization~\cite{sanchez}. Moreover, we provide the rationale why MPN-COV well addresses the two challenges of the covariance pooling.   Finally, DeepO$^{2}$P and B-CNN  have  not been evaluated on  challenging, large-scale ImageNet dataset.

\begin{figure}[tb]
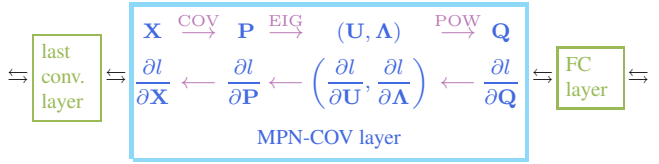

\footnotesize
\setlength\tabcolsep{0pt}
\renewcommand{\arraystretch}{2.0}
\setlength\arrayrulewidth{1.2pt}\arrayrulecolor{tabcolor1}
\begin{tabular}{c|ccccccc|c}
\cline{2-8}
\multirow{3}{*}{$\leftrightarrows$\,\textcolor{textcolor1}{\fbox{\parbox{0.27in}{\textcolor{textcolor1}{last conv. layer}}}}\,$\leftrightarrows$\,}
& \textcolor{textcolor3}{$\mathbf{X}$ }& \textcolor{textcolor2}{$\stackrel{\mathrm{COV}}{\longrightarrow}$ }   & \textcolor{textcolor3}{$\mathbf{P}$}& \textcolor{textcolor2}{$\stackrel{\mathrm{EIG}}{\longrightarrow}$ }& \textcolor{textcolor3}{$(\mathbf{U},\boldsymbol{\Lambda})$ }& \textcolor{textcolor2}{$\stackrel{\mathrm{POW}}{\longrightarrow}$} & \textcolor{textcolor3}{$\mathbf{Q}$ }& \multirow{3}{*}{\,$\leftrightarrows$\,\textcolor{textcolor1}{\fbox{\parbox{0.26in}{\textcolor{textcolor1}{FC layer}}}}\,$\leftrightarrows$}\\
&\textcolor{textcolor3}{$\dfrac{\partial l}{\partial \mathbf{X}}$ } & \textcolor{textcolor2}{$\longleftarrow$ } & \textcolor{textcolor3}{$\dfrac{\partial l}{\partial \mathbf{P}}$ } & \textcolor{textcolor2}{$\longleftarrow$ }& \textcolor{textcolor3}{$\bigg(\dfrac{\partial l}{\partial \mathbf{U}},\dfrac{\partial l}{\partial \boldsymbol{\Lambda}}\bigg)$ }&\textcolor{textcolor2}{ $\longleftarrow$} & \textcolor{textcolor3}{$\dfrac{\partial l}{\partial \mathbf{Q}}$ } & \\
&\multicolumn{7}{c|}{\textcolor{layercolor1}{MPN-COV layer}}&\\
\cline{2-8}
\end{tabular}
\arrayrulecolor{black}
\vspace{2.0mm}
\caption{Illustration of forward  and backward  propagations of ConvNets with MPN-COV. The proposed MPN-COV as a layer is inserted between the last conv. layer and FC layer, trainable end-to-end.  See text for notations and in-depth description.}
\label{figure:overview}
\end{figure}

\section{The Proposed MPN-COV}\label{section:proposed-method-MPN-COV}

For an input image, MPN-COV produces a  normalized covariance matrix as a representation, which characterizes the correlations of feature channels and actually designates the shape of feature distribution. Fig.~\ref{figure:overview} illustrates the forward and backward propagations of MPN-COV. Given the responses $\mathbf{X}$  of the last conv. layer as features, we first compute the sample covariance  matrix $\mathbf{P}$ of $\mathbf{X}$. Then we perform eigenvalue decomposition (EIG) of $\mathbf{P}$ to obtain the orthogonal matrix $\mathbf{U}$ and diagonal matrix $\boldsymbol{\Lambda}$, through which the matrix power $\mathbf{Q}\stackrel{\vartriangle}{=}\mathbf{P}^{\alpha}$ can be transformed to the power of eigenvalues of $\mathbf{P}$. Finally, $\mathbf{Q}$ will be inputted to the subsequent, top FC layer.  Accordingly, in backward pass, given  the partial derivative $\frac{\partial l}{\partial \mathbf{Q}}$ of loss function $l$ with respect to $\mathbf{Q}$ propagated from the top FC layer,  we need to  compute in reverse order the associated partial derivatives.

\subsection{Forward Propagation}\label{subsection:forward-propgation}

Let $\mathbf{X}\in \mathbb{R}^{d\times N}$ be a matrix whose columns consist of a sample of $N$ features of dimension $d$. The sample covariance matrix  $\mathbf{P}$ of $\mathbf{X}$ is computed as
\begin{align}\label{equ:compute-cov}
\mathbf{X}\mapsto \mathbf{P}, \quad \mathbf{P}=\mathbf{X}\bar{\mathbf{I}}\mathbf{X}^{T}.
\end{align}
Here  $\bar{\mathbf{I}}=\frac{1}{N}(\mathbf{I}-\frac{1}{N}\mathbf{1}\mathbf{1}^{T})$, where  $\mathbf{I}$ is the $N\times N$ identity matrix, $\mathbf{1}=[1,\ldots,1]^{T}$ is a $N-$dimensional  vector, and  $T$ denotes the matrix transpose. 
The sample covariance matrix $\mathbf{P}$ is symmetric positive semi-definite, which has eigenvalue decomposition as  follows: 
\begin{align}\label{equ:eig-cov}
\mathbf{P}\mapsto (\mathbf{U},\boldsymbol{\Lambda}), \quad \mathbf{P}=\mathbf{U}\boldsymbol{\Lambda}\mathbf{U}^{T},
\end{align}
where $\boldsymbol{\Lambda}=\mathrm{diag}(\lambda_{1},\ldots,\lambda_{d})$ is a diagonal matrix and $\lambda_{i}, i=1,\ldots, d$ are eigenvalues arranged in non-increasing order; $\mathbf{U}=[\mathbf{u}_{1},\ldots,\mathbf{u}_{d}]$ is an orthogonal matrix whose column $\mathbf{u}_{i}$ is  the eigenvector corresponding to $\lambda_{i}$.
Through EIG we can convert matrix power to the power of eigenvalues. Hence, we have  
\begin{align}\label{equ:normlization-cov}
(\mathbf{U},\boldsymbol{\Lambda})\mapsto \mathbf{Q},\;\;\mathbf{Q}\stackrel{\vartriangle}{=}\mathbf{P}^{\alpha}=\mathbf{U}\mathbf{F}(\boldsymbol{\Lambda})\mathbf{U}^{T}. 
\end{align}
Here $\alpha$ is a positive real number and $\mathbf{F}(\boldsymbol{\Lambda})=\mathrm{diag}(f(\lambda_{1}),\ldots,f(\lambda_{d}))$, where $f(\lambda_{i})$ is the power of eigenvalues
\begin{align}\label{equ:power-norm}
f(\lambda_{i})=\lambda_{i}^{\alpha},\quad \text{for MPN}.
\end{align}

Inspired by the element-wise power normalization technique~\cite{sanchez}, we can further perform, right after MPN, normalization by matrix $\ell_{2}-$norm (M-$\ell_{2}$) or by  matrix Frobenius norm (M-Fro). The matrix $\ell_{2}-$norm (also known as the \textit{spectral norm})  of a matrix  $\mathbf{P}$, denoted by $\|\mathbf{P}\|_{2}$, is defined as the largest singular value of $\mathbf{P}$, which equals the largest eigenvalue if $\mathbf{P}$ is a covariance matrix. The matrix Frobenius norm of $\mathbf{P}$ can be defined in various ways such as $\|\mathbf{P}\|_{F}=(\mathrm{tr}(\mathbf{P}^{T}\mathbf{P}))^{\frac{1}{2}}=(\sum_{i}\lambda_{i}^{2})^{\frac{1}{2}}$, where $\lambda_{i}$ are singular values of $\mathbf{P}$. As such, we have 
\begin{align}\label{equ:MPN-l2-or-fro}
\renewcommand*{\arraystretch}{1.2}
f(\lambda_{i})=\left\{ \begin{matrix}
{\lambda_{i}^{\alpha}}\Big/{\lambda_{1}^{\alpha}} & \text{for MPN+M-$\ell_{2}$} \\
{\lambda_{i}^{\alpha}}\Big/(\sum_{k} \lambda_{k}^{2\alpha})^{\frac{1}{2}} & \text{for MPN+M-Fro}
\end{matrix}\right.
\end{align}
Note that, in (\ref{equ:MPN-l2-or-fro}),  when $\alpha=1$ the first and second identities reduce to separate M-$\ell_{2}$ and M-Fro normalizations, respectively.

\subsection{Backward Propagation}\label{subsection:backward-propgation}

We use the methodology of matrix backpropagation, formulated in~\cite{Ionescu_2015_ICCV,IonescuVS15},  to compute the partial derivative of loss function $l$ with respect to the input matrix of some layer. It is built on  the theory of matrix calculus, enabling inclusion of structured, nonlinear matrix functions in neural networks while considering  the invariants involved such as orthogonality, diagonality and symmetry. 

Let us  consider derivation of $\frac{\partial l}{\partial \mathbf{U}}$ and $\frac{\partial l}{\partial \boldsymbol{\Lambda}}$, given $\frac{\partial l}{\partial \mathbf{Q}}$ propagated from the top FC layer. The expression of the chain rule is 
\begin{align}\label{equ:chain-rule}
\mathrm{tr}\Big(\Big(\frac{\partial l}{\partial \mathbf{U}}\Big)^{T}\mathrm{d}\mathbf{U}+\Big(\frac{\partial l}{\partial \boldsymbol{\Lambda}}\Big)^{T}\mathrm{d}\boldsymbol{\Lambda}\Big)=\mathrm{tr}\Big(\Big(\frac{\partial l}{\partial \mathbf{Q}}\Big)^{T}\mathrm{d}\mathbf{Q}\Big),
\end{align}
where $\mathrm{d}\mathbf{Q}$ denotes variation of matrix $\mathbf{Q}$. From Eq.~ (\ref{equ:normlization-cov}), we have $\mathrm{d}\mathbf{Q}=\mathrm{d}\mathbf{U}\mathbf{F}\mathbf{U}^{T}+\mathbf{U}\mathrm{d}\mathbf{F}\mathbf{U}^{T}+\mathbf{U}\mathbf{F}\mathrm{d}\mathbf{U}^{T}$.  We note that  $\mathrm{d}\mathbf{F}=\mathrm{diag}\big(\alpha \lambda_{1}^{\alpha-1}, \ldots, \alpha \lambda_{d}^{\alpha-1}\big)\mathrm{d}\boldsymbol{\Lambda}$. After some arrangements, we  obtain   
\begin{align}\label{equ:backward_step-norm}
\dfrac{\partial l}{\partial \mathbf{U}}&= \Big(\dfrac{\partial l}{\partial \mathbf{Q}}+\Big(\dfrac{\partial l}{\partial \mathbf{Q}}\Big)^{T}\Big)\mathbf{U}\mathbf{F}\\
\dfrac{\partial l}{\partial \boldsymbol{\Lambda}}&=\alpha\Big(\mathrm{diag}\Big( \lambda_{1}^{\alpha-1}, \ldots,  \lambda_{d}^{\alpha-1}\Big)\mathbf{U}^{T}\dfrac{\partial l}{\partial \mathbf{Q}}\mathbf{U}\Big)_{\mathrm{diag}},\nonumber
\end{align}
where $\mathbf{A}_{\mathrm{diag}}$ denotes the operation preserving the diagonal entries of $\mathbf{A}$ while setting all non-diagonal entries to zero.
For  MPN+M-$\ell_{2}$ and MPN+M-Fro, $\frac{\partial l}{\partial \boldsymbol{\Lambda}}$ takes respectively the following forms: 
\begin{align}\label{equ:backward_step-power_l2}
\dfrac{\partial l}{\partial \boldsymbol{\Lambda}}=&\dfrac{\alpha}{\lambda_{1}^{\alpha}}\Big(\mathrm{diag}\Big({\lambda_{1}^{\alpha-1}}, \ldots, {\lambda_{d}^{\alpha-1}}\Big)\mathbf{U}^{T}\dfrac{\partial l}{\partial \mathbf{Q}}\mathbf{U}\Big)_{\mathrm{diag}}\\
&- \mathrm{diag}\bigg(\dfrac{\alpha}{\lambda_{1}}\mathrm{tr}\Big(\mathbf{Q}\dfrac{\partial l}{\partial \mathbf{Q}}\Big), 0, \ldots, 0\bigg)\nonumber
\end{align}
and
\begin{align}\label{equ:backward_step-power_fro}
\dfrac{\partial l}{\partial \boldsymbol{\Lambda}}=&\dfrac{\alpha}{\sqrt{\sum\nolimits_{k} \lambda_{k}^{2\alpha}}}\Big(\mathrm{diag}\Big({ \lambda_{1}^{\alpha-1}}, \ldots, { \lambda_{d}^{\alpha-1}}\Big)\mathbf{U}^{T}\dfrac{\partial l}{\partial \mathbf{Q}}\mathbf{U}\Big)_{\mathrm{diag}}\nonumber\\
-&\dfrac{\alpha}{{\sum\nolimits_{k} \lambda_{k}^{2\alpha}}}\mathrm{tr}\Big(\mathbf{Q}\dfrac{\partial l}{\partial \mathbf{Q}}\Big)\mathrm{diag}\Big({ \lambda_{1}^{2\alpha-1}}, \ldots, {\lambda_{d}^{2\alpha-1}}\Big).
\end{align}

Next, given  $\frac{\partial l}{\partial \mathbf{U}}$ and $\frac{\partial l}{\partial \boldsymbol{\Lambda}}$, let us compute $\frac{\partial l}{\partial \mathbf{P}}$  associated with EIG (\ref{equ:eig-cov}). The chain rule is $\mathrm{tr}((\frac{\partial l}{\partial \mathbf{P}})^{T}\mathrm{d}\mathbf{P})=\mathrm{tr}((\frac{\partial l}{\partial \mathbf{U}})^{T}\mathrm{d}\mathbf{U}+(\frac{\partial l}{\partial\boldsymbol{\Lambda} })^{T}\mathrm{d}\boldsymbol{\Lambda})$. Note that $\mathbf{U}$ should satisfy the orthogonal constraint. After some arrangements, we have 
\begin{align}\label{equ:eigendecomposiiton-backward}
\dfrac{\partial l}{\partial \mathbf{P}}=\mathbf{U}\Big(\Big(\mathbf{K}^{T}\circ \Big(\mathbf{U}^{T}\dfrac{\partial l}{\partial \mathbf{U}}\Big)\Big)+\Big(\dfrac{\partial l}{\boldsymbol{\partial \Lambda}}\Big)_{\mathrm{diag}}\Big)\mathbf{U}^{T},
\end{align}
where $\circ$ denotes matrix Kronecker product. 
The  matrix $\mathbf{K}=\{K_{ij}\}$ where $K_{ij}=1/(\lambda_{i}-\lambda_{j})$ if $i\neq j$ and $K_{ij}=0$ otherwise. We refer readers to ~\cite[Proposition 2]{IonescuVS15} for in-depth derivation of Eq.~(\ref{equ:eigendecomposiiton-backward}). 

Finally, given $\frac{\partial l}{\partial \mathbf{P}}$, we derive the gradient of the loss function with respect to the input matrix $\mathbf{X}$, which takes the following form:
\begin{align}\label{equ:compute-cov-backward}
\dfrac{\partial l}{\partial \mathbf{X}}=\bar{\mathbf{I}}\mathbf{X}\bigg(\dfrac{\partial l}{\partial \mathbf{P}}+\bigg(\dfrac{\partial l}{\partial \mathbf{P}}\bigg)^{T}\bigg).
\end{align}

\section{The Mechanism Underlying MPN-COV}\label{section: why-proposed-works}

This section explains the mechanism underlying MPN-COV. We provide interpretations from the statistical and geometric points of view, and make qualitative analysis from computational perspective.

\subsection{MPN-COV Amounts to Robust Covariance Estimation}\label{subsection:robust-estimator}

The sample covariance  amounts to the solution to the Maximum Likelihood Estimation (MLE)  of normally distributed random vectors. Though MLE is widely used to estimate covariances, it is well known that it performs poorly when the sample of data is of large dimension with small size~\cite{ChenWEH10,icml2014c2_yangd14}. This is just what our covariance pooling faces: in most state-of-the-art ConvNets~\cite{Simonyan15,icml2015_ioffe15,He_2016_CVPR}, the last convolutional layer outputs  less than 200 features of dimension larger than 512, and so the sample covariances are always rank-deficient, rendering robust estimation  critical.

The robust estimation of large-dimensional covariances with small  sample size has been of great interest in statistics~\cite{Stein1986}, signal processing~\cite{ChenWEH10} and biology~\cite{icml2014c2_yangd14}.  
Stein~\cite{Stein1986} for the first time proposes the shrinkage principle for eigenvalues of sample covariances. Ledioit and Wolf~\cite{ledoit2004wellconditioned} has shown that the largest eigenvalues are systematically biased upwards while the smallest ones are biased downwards, and thus introduced the optimal linear shrinkage estimator, where the estimated covariance matrix $\mathbf{Q}$  is a linear combination of the sample covariance $\mathbf{P}$  with the identity matrix (i.e., $\mathbf{Q}=\alpha_{1}\mathbf{P}+\alpha_{2}\mathbf{I}$). This method with $\alpha_i$ decided by cross-validation  is widely used to counteract the ill-conditioning of  covariance matrices. 
Our MPN-COV  closely conforms to the shrinkage principle~\cite{Stein1986,ledoit2004wellconditioned}, i.e., shrinking the largest sample eigenvalues and stretching the smallest ones, as will be shown later in Sec.~\ref{subsection:computational-perspective}. It only depends on the sample covariance, delivering an individualized shrinkage intensity to each eigenvalue. 

A number of researchers propose various regularized MLE methods for robust covariance estimation (see~\cite{icml2014c2_yangd14} and references therein). An important conclusion we can draw is that  MPN-COV can be deemed a robust covariance estimator, explicitly derived from a regularized MLE called vN-MLE, according to  our previous  work~\cite{Wang_2016_CVPR}. Specifically, we have
\begin{proposition}\label{propostion-robust-estimation}
MPN-COV with $\alpha=\frac{1}{2}$ is the unique solution to the regularized MLE of covariance matrix, i.e.,
\begin{align}\label{equ:robust-estimation}
\mathbf{P}^{\frac{1}{2}}=\arg\min_{\boldsymbol\Sigma} \log|{\boldsymbol\Sigma}| + \mathrm{tr}({\boldsymbol\Sigma}^{-1}{\mathbf{P}}) +  D_{\mathrm{vN}}(\mathbf{I},{\mathbf{\Sigma}}),
\end{align}
where $\boldsymbol\Sigma$ is constrained to be positive semidefinite, and  $D_{\text{vN}}(\mathbf{A},\mathbf{B})=\mathrm{tr}(\mathbf{A}(\log(\mathbf{A})-\log(\mathbf{B}))-\mathbf{A}+\mathbf{B})$ is matrix von Neumann divergence.
\end{proposition}
Proposition~\ref{propostion-robust-estimation} follows immediately  by setting to one the regularizing parameter in~\cite[Theorem 1]{Wang_2016_CVPR}. Note that the classical MLE only includes the first two terms on the right-hand side of Eq. (\ref{equ:robust-estimation}), while the robust vN-MLE estimator introduces the third term, constraining  the covariance matrix be similar to the identity matrix.  It has been shown~\cite{Wang_2016_CVPR} that the vN-MLE  outperforms other shrinkage methods~\cite{Stein1986,ledoit2004wellconditioned,ChenWEH10} and  regularized MLE method~\cite{icml2014c2_yangd14}.

\subsection{MPN-COV Approximately Exploits Riemannian Geometry}\label{subsection:exploit-geometry}

As the space of $d\times d$ covariance matrices, denoted by $Sym^{+}_{d}$, is a Riemannian manifold,  it is appropriate to consider the geometrical structure when operating  on this manifold. There are mainly two kinds of  Riemannian metrics, i.e., the affine Riemannian metric~\cite{AffineMetricsIJCV06} and the Log-E metric~\cite{LogMetricsSIAM06}. The former metric is affine-invariant, but is computationally inefficient and is coupled, not scalable to large-scale setting. In contrast, the most often used Log-E metric is similarity-invariant, efficient to compute and scalable to large-scale problems  as it is a decoupled metric. 

The metric for MPN-COV corresponds to the Power Euclidean (Pow-E) metric~\cite{10.2307/30242879}. It has close connection with the Log-E metric, as presented in the following proposition: 

\begin{proposition}\label{proposition-power-Euclidean}
For any two covariance matrices $\mathbf{P}$ and $\widetilde{\mathbf{P}}$, the limit of the Pow-E metric 
$d_{\alpha}(\mathbf{P},\widetilde{\mathbf{P}})=\frac{1}{\alpha}\big\|\mathbf{P}^{\alpha}-\widetilde{\mathbf{P}}^{\alpha}\big\|_{F}$
as $\alpha>0$ approaches 0 equals the Log-E metric, i.e., 
$\lim\limits_{\alpha \rightarrow 0}d_{\alpha}(\mathbf{P},\widetilde{\mathbf{P}})=\big\|\log(\mathbf{P})-\log(\widetilde{\mathbf{P}})\big\|_{F}$.
\end{proposition}
This conclusion was first mentioned in~\cite{10.2307/30242879} but without proof. Here we briefly prove this claim. Note that $d_{\alpha}(\mathbf{P},\widetilde{\mathbf{P}})=\big\|\frac{1}{\alpha}(\mathbf{P}^{\alpha}-\mathbf{I})-\frac{1}{\alpha}(\widetilde{\mathbf{P}}^{\alpha}-\mathbf{I})\big\|_{F}$. 
Based on the eigenvalue decomposition of  $\mathbf{P}$  we have  $\frac{1}{\alpha}(\mathbf{P}^{\alpha}-\mathbf{I})=\mathbf{U}\mathrm{diag}(\frac{\lambda_{1}^{\alpha}-1}{\alpha},\ldots,\frac{\lambda_{n}^{\alpha}-1}{\alpha})\mathbf{U}^{T}$.  The identity about the limit in Proposition~\ref{proposition-power-Euclidean} follows immediately by recalling  $\lim_{\alpha\rightarrow 0}\frac{\lambda^{\alpha}-1}{\alpha}=\log(\lambda)$.

\begin{figure}[tb]
\centering
\subfigure[Eigenvalue histogram and  normalization functions]{\label{subfigure:hist-and-norm-functions}
\begin{minipage}[b]{0.48\linewidth}
\centering
\includegraphics[width=1.0\textwidth]{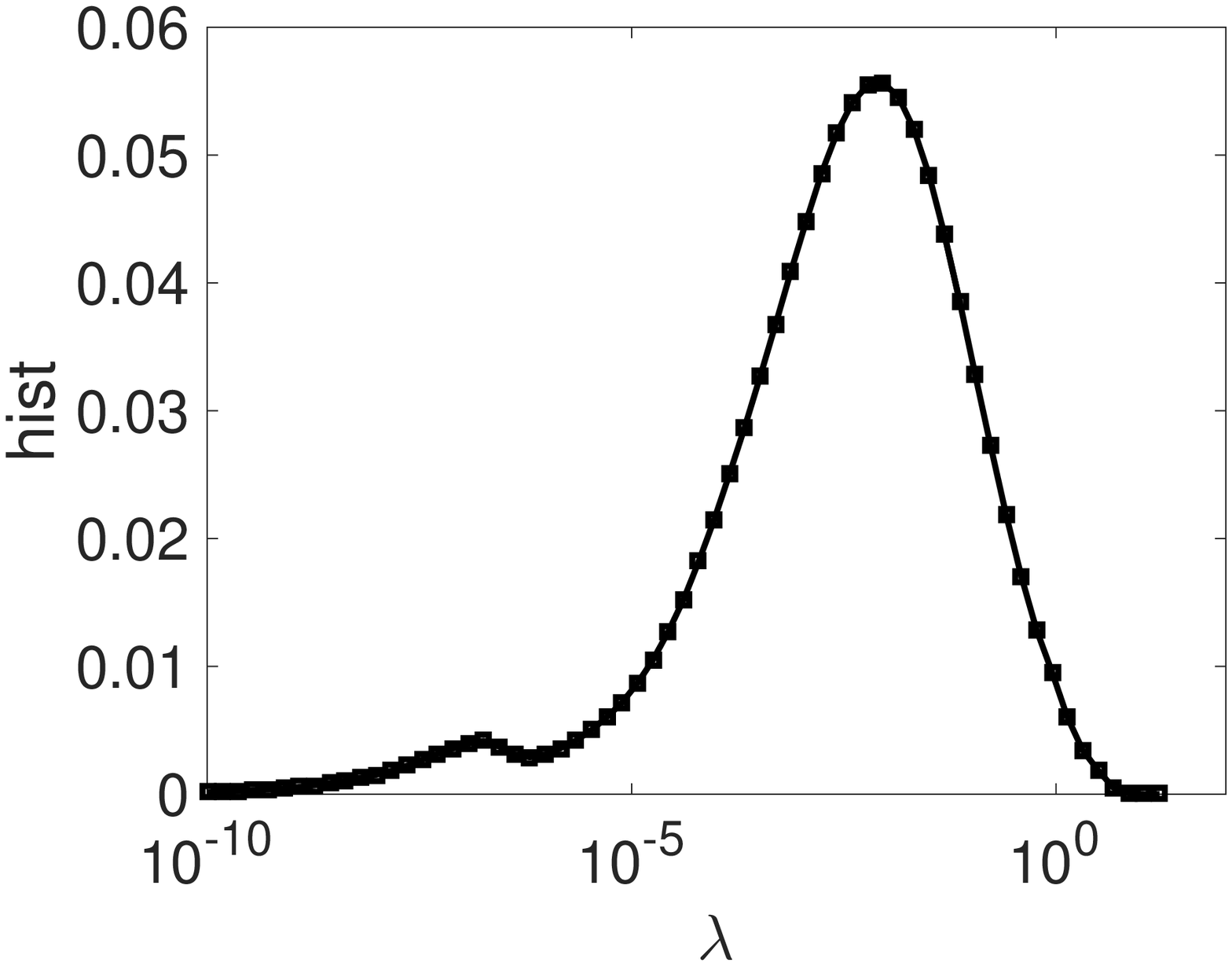}
\end{minipage}
\begin{minipage}[b]{0.48\linewidth}
\centering
\includegraphics[width=1.0\textwidth]{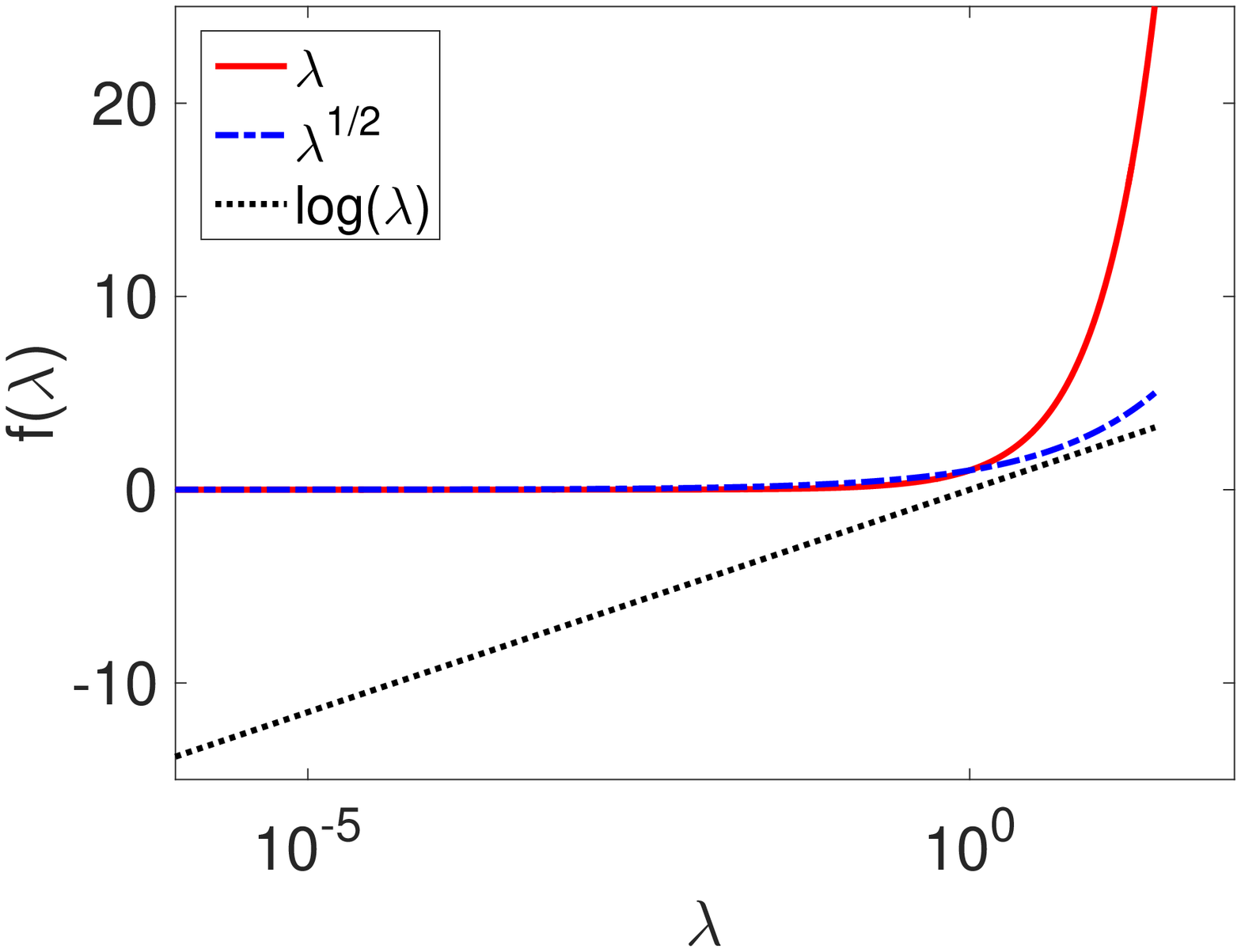}
\end{minipage}}

\subfigure[$\lambda^{\frac{1}{2}}$ and its derivative zoomed on ${[10^{-5},1]}$]{\label{subfigure:sqrt-and-derivative}
\begin{minipage}[b]{0.48\linewidth}
\centering
\includegraphics[width=1.0\textwidth]{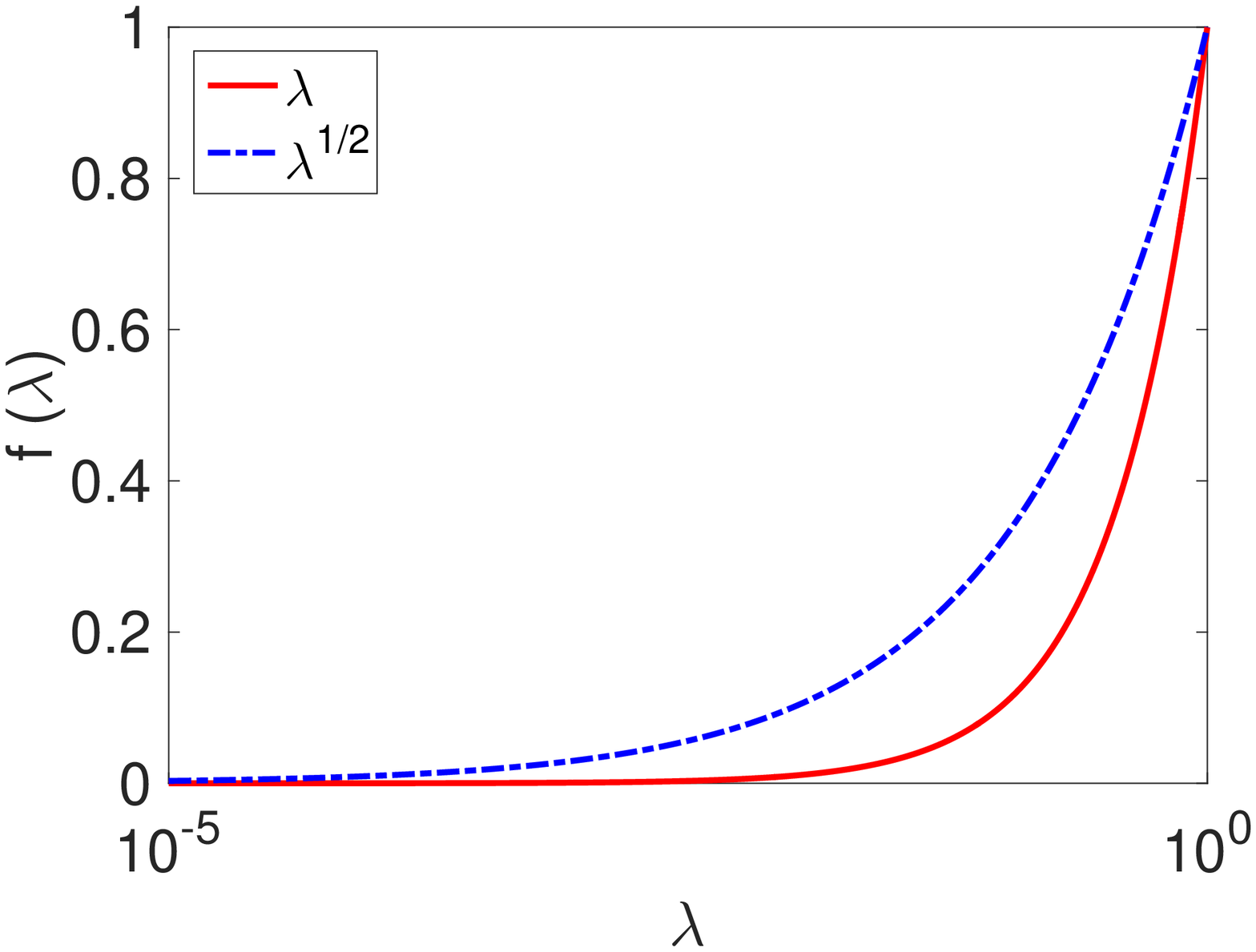}
\end{minipage}
\begin{minipage}[b]{0.48\linewidth}
\centering
\includegraphics[width=1.0\textwidth]{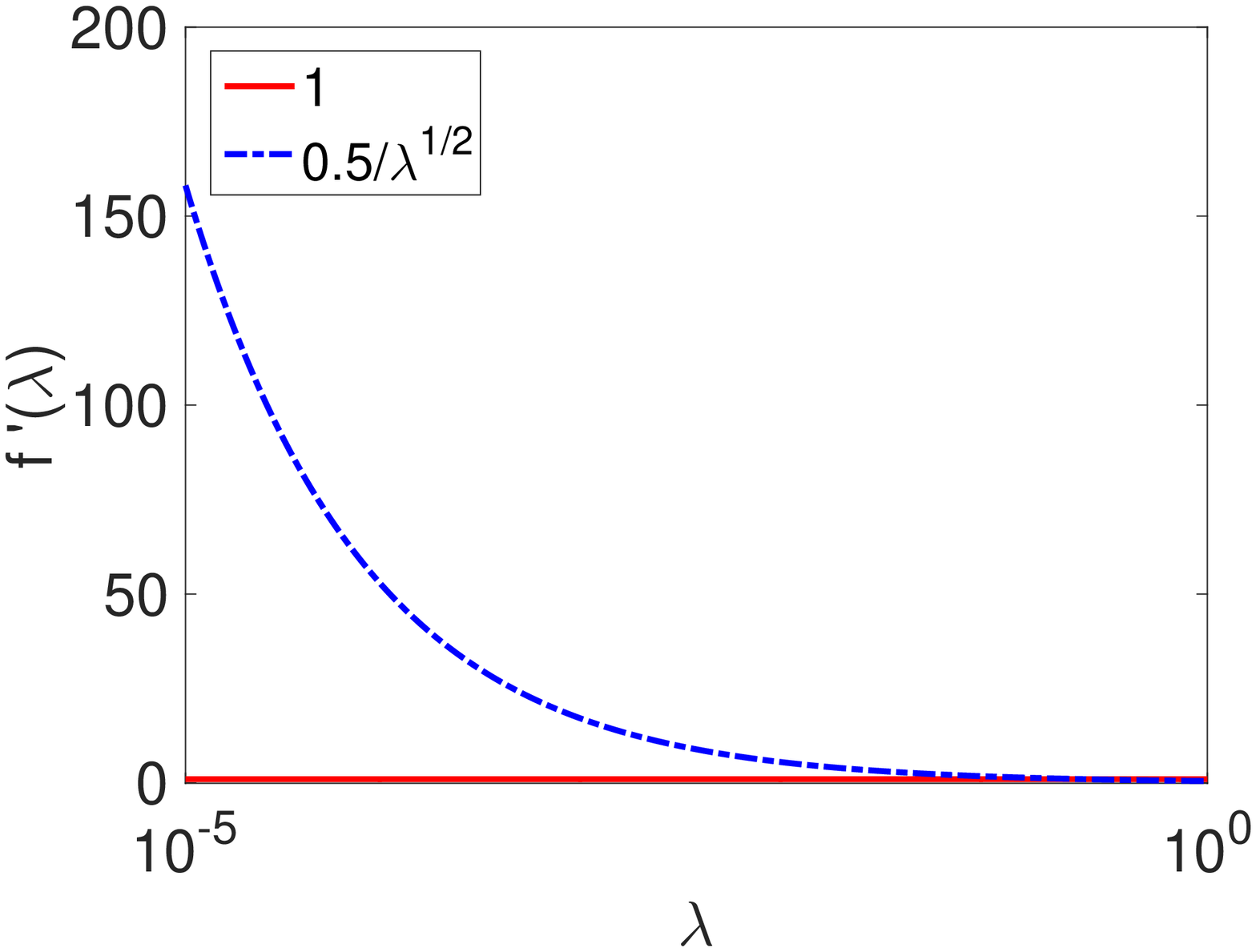}
\end{minipage}}

\subfigure[$\log(\lambda)$ and its derivative zoomed on ${[10^{-5},1]}$]{\label{subfigure:log-and-derivate}
\begin{minipage}[b]{0.48\linewidth}
\centering
\includegraphics[width=1.0\textwidth]{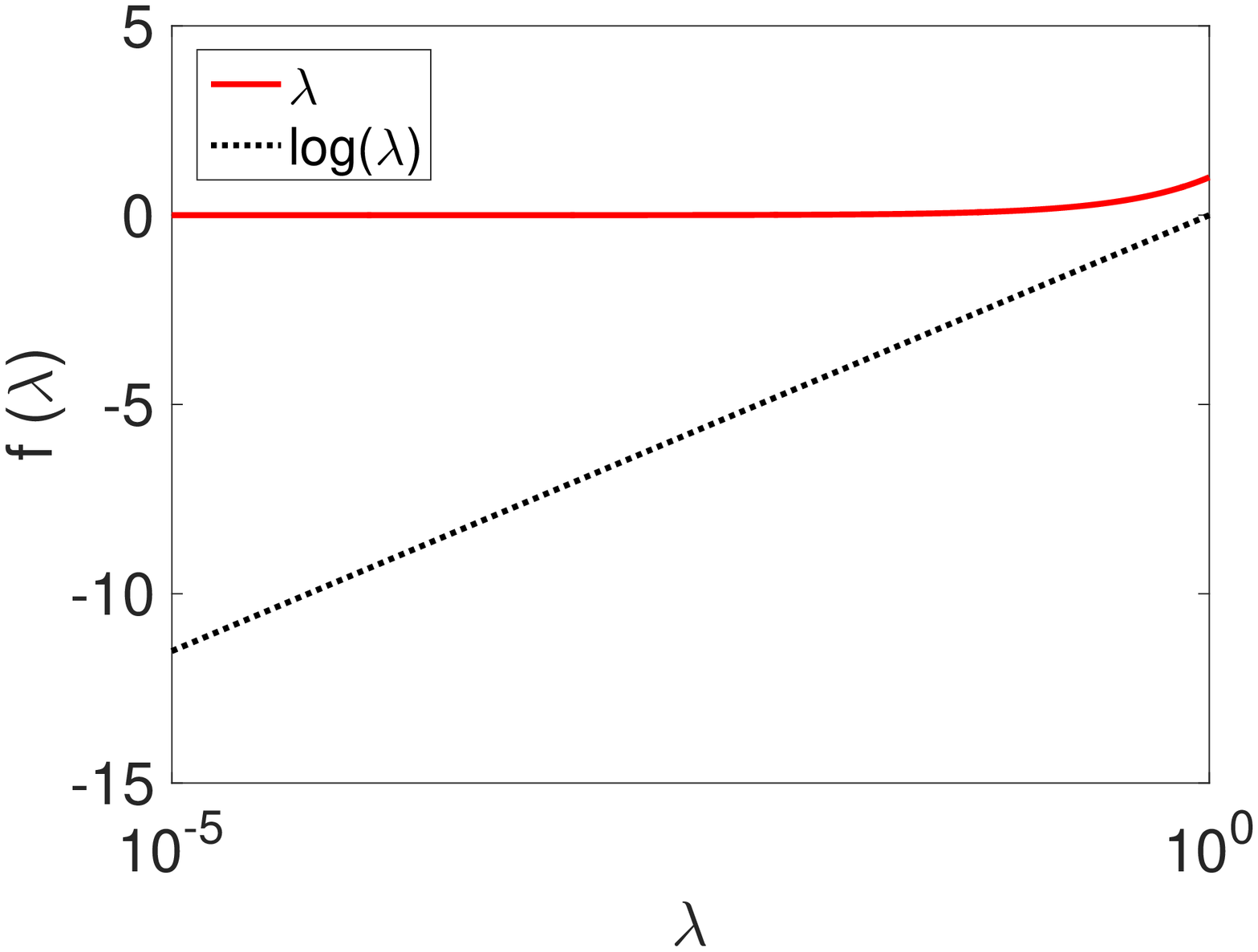}
\end{minipage}
\begin{minipage}[b]{0.48\linewidth}
\centering
\includegraphics[width=1.0\textwidth]{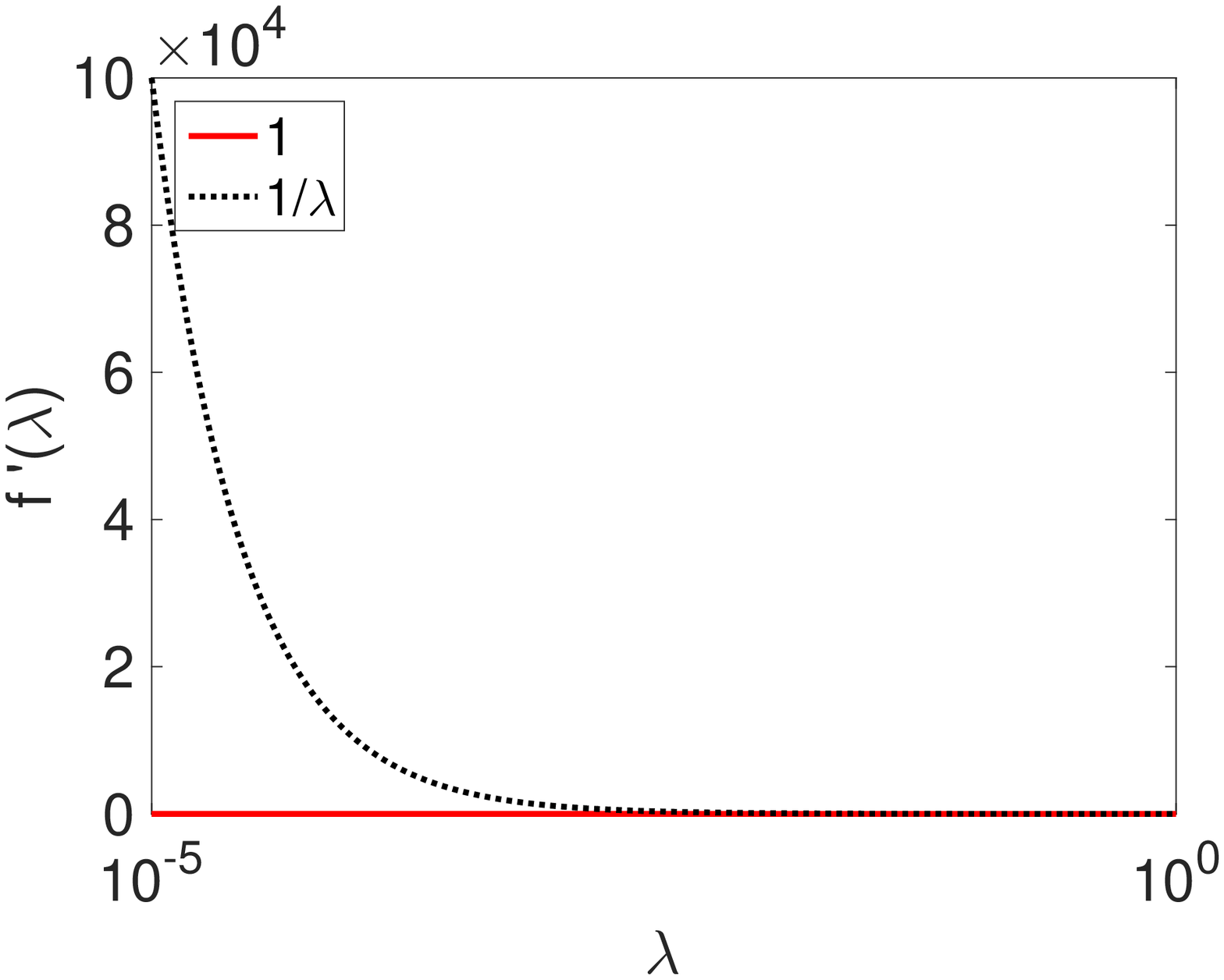}
\end{minipage}}
\caption{Illustration of empirical distribution of eigenvalues and  normalization functions.  The identity  $f(\lambda)=\lambda$ (no normalization) and its derivative are also plotted for reference. $\lambda^{\frac{1}{2}}$ conforms to the general shrinkage principle as suggested in~\cite{Stein1986,ledoit2004wellconditioned}, which shrinks the largest eigenvalues and stretches the smallest ones, while preserving the order of eigenvalue significances. In contrast, $\log(\lambda)$ over-stretches the smallest eigenvalues, reversing the order of significance such that, after normalization, the smallest eigenvalues play more crucial roles than the largest ones.}
\label{figure:hist-and-normalization functions}
\end{figure}

Hence, the proposed MPN-COV can be viewed as approximately exploiting the Riemannian geometry of $Sym^{+}_{d}$. It might seem that the Log-E metric  is a better option than the Pow-E metric, since the former measures the true geodesic distance but the latter one only measures it approximately. We argue that  this is not the case of our problem for two reasons. First, the Log-E metric requires the eigenvalues involved to be \emph{strictly positive}~\cite{LogMetricsSIAM06,doi:10.1080/02664763.2015.1080671} while the Pow-E metric allows \emph{non-negative} eigenvalues~\cite{10.2307/30242879,doi:10.1080/02664763.2015.1080671}. For Log-E  the  common method is to add a small positive number $\epsilon$  to  eigenvalues  for improving numerical stability. Although  $\epsilon$ can be decided by cross-validation, it is difficult to seek a particular $\epsilon$ well suitable for a huge number of images. For example, \cite{Ionescu_2015_ICCV} suggest $\epsilon=10^{-3}$, which will  smooth out eigenvalues less than $10^{-3}$. Above all, the distributions of high-level, convolutional features are such that the logarithm  brings side effect, which will be qualitatively analyzed  in the next subsection. We will also  quantitatively compare the two metrics by experiments in Sec.~\ref{subsection:Experiment-evaluation}.

\subsection{Qualitative Analysis}\label{subsection:computational-perspective}

This section qualitatively analyzes, from the computational perspective,  the impact of matrix power  and  logarithm  on the eigenvalues of sample covariances. The matrix logarithm can be regarded as a kind of  normalization, nonlinearly applied to the eigenvalues:
$\mathbf{Q}\stackrel{\vartriangle}{=}\log(\mathbf{P})=\mathbf{U}\mathrm{diag}(\log(\lambda_{1}),\ldots,\log(\lambda_{d}))\mathbf{U}^{T}$.
Below we will concentrate on power function $f(\lambda)=\lambda^{\frac{1}{2}}$  and logarithm  $f(\lambda)=\log(\lambda)$. 

We first examine the empirical distribution of eigenvalues of sample covariances. We randomly select 300,000 images from the training set of ImageNet 2012. For each image, we extract the output of the 5th conv. (Conv5) layer (with ReLU)  using \href{http://www.vlfeat.org/matconvnet/models/imagenet-matconvnet-alex.mat}{AlexNet} model pretrained on ImageNet 2012, estimate the sample covariance  $\mathbf{P}$, and then compute its eigenvalues using EIG in single-precision floating-point format. For a training image of $227\times 227$, Conv5 outputs  $13\times 13$ features with 256 channels, reshaped to a  matrix $\mathbf{X}\in \mathds{R}^{256\times 169}$. As the rank of $\mathbf{P}$ is less than 169,  $\mathbf{P}$ has less than 169 non-zeros eigenvalues. We mention that very small eigenvalues obtained by EIG  may be inaccurate due to machine precision.  The histogram of eigenvalues is shown in Fig.~\ref{subfigure:hist-and-norm-functions}(left), where zero eigenvalues are excluded for better view.  Fig.~\ref{subfigure:hist-and-norm-functions}(right) shows the two normalization functions  over $[10^{-5}, 10]$. The graphs of $\lambda^{\frac{1}{2}}$ \&  its derivative and $\log(\lambda)$ \& its derivative, both zoomed on $[10^{-5},1]$, are shown in Fig.~\ref{subfigure:sqrt-and-derivative} and Fig.~\ref{subfigure:log-and-derivate}, respectively. 

The function $\log(\lambda)$ considerably changes the eigenvalue magnitudes, reversing the order of eigenvalue significances, e.g., a significant eigenvalue $\lambda=50\mapsto \log(\lambda)\approx 3.9$ but an insignificant one $\lambda=10^{-3}\mapsto \log(\lambda)\approx-6.9$. From the forward propagation formula $\mathbf{P}=\sum_{i}\lambda_{i}\mathbf{u}_{i}\mathbf{u}_{i}^{T}\mapsto \mathbf{Q}=\sum_{i}\log(\lambda_{i})\mathbf{u}_{i}\mathbf{u}_{i}^{T}$
it can be seen that the smallest eigenvalues will  play  more crucial roles than the largest ones.  This effect is also obvious if we consider the backprogation formula for the gradient $\frac{\partial l}{\partial \lambda_{i}}$ before and after normalization, i.e., $\mathbf{u}_{i}^{T}\frac{\partial l}{\partial \mathbf{Q}}\mathbf{u}_{i} \mapsto \frac{1}{\lambda_{i}}\mathbf{u}_{i}^{T}\frac{\partial l}{\partial \mathbf{Q}}\mathbf{u}_{i},i=1,\ldots, d$. For example,  the derivative of $\log(\lambda)$ at  $\lambda=10^{-3}$ is $10^3$ but  at  $\lambda=50$ is $2\times 10^{-2}$. Since significant eigenvalues are generally more important in that they capture the statistics of principal directions along which the feature variances are larger,  matrix logarithm will deteriorate the covariance representations. 

Now let us consider $\lambda^{\frac{1}{2}}$. It nonlinearly shrinks the eigenvalues larger than one, and the larger, the more shrunk, while stretching those less than one, and the smaller, the more stretched. This kind of normalization conforms to the general shrinkage principle as suggested in~\cite{Stein1986,ledoit2004wellconditioned}. Contrary to $\log(\lambda)$, it does not change the order of eigenvalue significances-- significant (resp. insignificant) eigenvalues maintain significant (resp. insignificant). For example,  $\lambda=50\mapsto \lambda^{\frac{1}{2}}\approx 7.1$ while  $\lambda=10^{-3}\mapsto \lambda^{\frac{1}{2}}\approx 0.032$. From the forward propagation formula  $\mathbf{S}=\sum_{i}\lambda_{i}\mathbf{u}_{i}\mathbf{u}_{i}^{T}\mapsto\mathbf{Q}=\sum_{i}\lambda^{\frac{1}{2}}\mathbf{u}_{i}\mathbf{u}_{i}^{T}$,  we see that the order of amount of contributions made by individual eigenvalues  keep unchanged. Similar conclusion can be drawn if we consider the backpropagation formula of $\frac{\partial l}{\partial \lambda_{i}}$: $\mathbf{u}_{i}^{T}\frac{\partial l}{\partial \mathbf{Q}}\mathbf{u}_{i} \mapsto \frac{1}{2\sqrt{\lambda_{i}}}\mathbf{u}_{i}^{T}\frac{\partial l}{\partial \mathbf{Q}}\mathbf{u}_{i}$.

\section{Experiments}\label{section:Experiment}

We make experiments on  ImageNet 2012 classification dataset~\cite{Russakovsky2015}, which consists of 1,000 classes, including roughly 1.28 million training images, 50k validation images, and 100k testing ones. We do not adopt  extra training images. Following the common practice, we report top-1 and top-5 error rates on the validation set as measures of recognition performance.   We develop programs based on  MatConvNet~\cite{vedaldi15matconvnet} and Matlab 2015b under 64-bit Windows 7.0. The programs run on six workstations  each of which is equipped with a Intel i7-4790k@4.0Ghz CPU and 32G RAM. Two NVIDA Titan X with  12 GB memory and four NVIDA GTX 1080 with 8 GB  memory are used, one graphics card per workstation.  

\subsection{Implementation of  MPN-COV Networks}

To implement MPN-COV layer, we adopt the EIG algorithm  on CPU in single-precision floating-point format, as its GPU version provided on the  CUDA platform is several times slower. Except for EIG, all  other operations  in forward and backward propagations are performed on GPU.  Since  MPN-COV allows non-negative eigenvalues,  we  truncate to zeros the eigenvalues  smaller than $\mathrm{eps}(\lambda_{1})$, which  indicates  the positive  distance from  the maximum eigenvalue $\lambda_{1}$ to its next larger floating-point number. Our MPN-COV pooling replaces the common  first-order, max/average pooling  after the last conv. layer, producing a global,  $d(d+1)/2-$dimensional image representation by concatenation of the upper triangular part  of one covariance matrix. In state-of-the-art ConvNets, the feature dimension $d$ of the last conv. layer gets much larger. For such architectures, we add a $1\times 1$ conv. layer of 256 channels  after the last conv. layer, so that the dimension of features inputted to the MPN-COV layer is fixed to 256 (see Sec.~\ref{subsection-MPN-COV-VGG} and Sec.~\ref{subsection:MPN-COV-ResNet}). As such, we alleviate the problem of small sample of large-dimensional features while decreasing the computational cost of the MPN-COV layer.

We adopt the standard color jittering technique~\cite{nips2012cnn} for training set augmentation. For AlexNet~\cite{nips2012cnn} and VGG-M~\cite{DBLP:conf/bmvc/ChatfieldSVZ14}, we follow the default setting in MatConvNet~\cite{vedaldi15matconvnet} where each training image is rescaled such that  its shorter side is of 256 pixels. For  VGG-16~\cite{Simonyan15} and ResNet~\cite{He_2016_CVPR}, following~\cite{Simonyan15}, we rescale isotropically each training image with shorter side randomly sampled on $[256, 512]$. Then,  we  sample a fixed size patch at random from the resized image or its mirror, and subtract the mean RGB value from each pixel. In testing stage, we first isotropically resize each test image with short side 256, then adopt the commonly used 1-crop prediction or 10-crop prediction for performance evaluation. Following~\cite{icml2015_ioffe15}, we adopt batch normalization right after every convolution and before ReLU and no drop out. 

We use mini-batch stochastic gradient descent  with momentum (set to 0.9 throughout the experiments) for training. For AlexNet, VGG-M and VGG-16, we set the weight decay to $5\times 10^{-4}$, and their  mini-batch sizes  are set to 128, 100 and 32, respectively. For training from scratch, the filter weights are initialized with a normal distribution $\mathcal{N}(0,0.01)$ with mean 0 and  variance 0.01 and the biases are initialized with zero~\cite{Simonyan15}; ConvNets are trained up to 20 epochs, where the learning rates  follow  exponential decay, changing from $10^{-1}$ to $10^{-4}$ and $10^{-1.2}$ to $10^{-5}$ for the ConvNets with  first-order pooling and those with MPN-COV pooling, respectively. For ResNets, following~\cite{He_2016_CVPR}, we use a weight decay of $10^{-4}$ and a mini-batch size of 256, and initialize the biases with zero and the filter weights with $\mathcal{N}(0,2/n)$,  where $n$ is the product of the size and \#channels of filters.  The ResNet-50 with MPN-COV  is trained up to 90 epochs with learning rate initialized to $10^{-1.2}$ and divided by 10 when the error plateaus.

\begin{figure}
\centering
\begin{minipage}[b]{0.65\linewidth}
\centering
\includegraphics[width=1.0\textwidth]{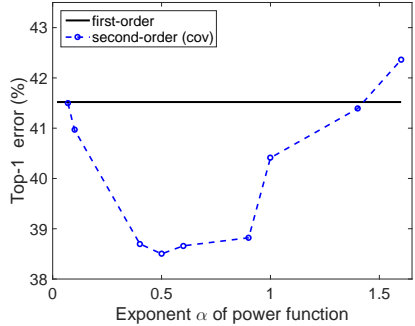}
\end{minipage}
\caption{Impact of  $\alpha$  on second-order cov pooling under AlextNet Architecture. Top-1 errors (\textit{1-crop})  are reported. The bold  line indicates the result of the AlexNet with first-order  pooling.}
\label{fig:impact_alpha}
\end{figure}

\subsection{Evaluation of MPN-COV Under AlexNet Architecture} \label{subsection:Experiment-evaluation}

In the first part of experiments, we evaluate MPN-COV by selecting  AlexNet architecture~\cite{nips2012cnn}, since it is shallower and  runs faster than its variants. As recently proposed deeper  ConvNets~\cite{Simonyan15,He_2016_CVPR}  follow the basic architecture of AlexNet, our analysis here  can extrapolate to them.  We study the impact of exponent $\alpha$ of power function and various matrix normalization methods  on cov pooling. We also compare with two existing end-to-end, second-order pooling methods, i.e.,   DeepO$^2$P~\cite{Ionescu_2015_ICCV} concerned with matrix normalization by logarithm and B-CNN~\cite{lin2015bilinear} which performs element-wise power normalization.

\begin{table}[tb]
\renewcommand{\baselinestretch}{1.3}
\footnotesize
\centering
\setlength\tabcolsep{5pt}
\subtable{
\begin{minipage}[t]{1.0\linewidth}
\centering
\begin{tabular}{lllllc}
\hline
method & MPN   & M-Fro & M-$\ell_2$ & init.  & top-1 err. \\
\hline
First-order (ours) & -- & -- & -- & Random & 41.52\\
\hline
\multirow{7}{*}{cov pool. (ours)}
& No    & No    & No   &  Random   &   40.41 \\
& Yes   & No    & No   &  Random   &   \textcolor{Blue}{\textbf{38.51}}  \\ 
& No    & Yes   & No   &  Random   &   39.87  \\
& No    & No    & Yes  &  Random   &  39.65\\
& Yes   & Yes   & No   &  Random   &  39.93\\
& Yes   & No    & Yes  &  Random   &  39.62\\
& Yes   & No    & No   &  Warm   &    {\textbf{37.35}}\\
\hline
\end{tabular}
\end{minipage}
 }
\renewcommand{\baselinestretch}{1.0}
\caption{Impact of various matrix normalizations under AlexNet architecture. We measure top-1  error (\%, \textit{1-crop})  for MPN, M-Fro and M-$\ell_{2}$ as well as combinations of them.}
\label{table:different-norm}

\vspace{3pt}
\setlength\tabcolsep{8pt}
\renewcommand{\baselinestretch}{1.1}
\small
\centering
\begin{minipage}[t]{1.0\linewidth}
\centering
\begin{tabular}{lll}
\hline
method & top-1 error   &  top-5 error \\
\hline
MPN-COV (AlexNet) (ours) & \textbf{33.84} & \textbf{14.01}\\
Krizhevsky et al.~\cite{nips2012cnn} & 40.7 & 18.2  \\
\hline
VGG-F~\cite{DBLP:conf/bmvc/ChatfieldSVZ14}   & 39.11 & 16.77  \\
\hline
\end{tabular}
\end{minipage}
\renewcommand{\baselinestretch}{1.0}
\caption{Error~(\%, \textit{10-crop}) comparison of MPN-COV (AlexNet)   with  ConvNets having similar architecture.}
\label{table:different-norm-comparision-related}
\end{table}

\vspace{3pt}\noindent\textbf{Impact of Exponent $\alpha$ of Power Function.}\;\; We first evaluate   covariance (cov) pooling against the exponent $\alpha$ of power function. Fig.~\ref{fig:impact_alpha} shows  top-1 errors versus  $\alpha$ using single-crop prediction. We first note that the \emph{plain cov pooling} ($\alpha=1$, no normalization) produces an error rate of 40.41\%, about  $1.1\%$ less than   first-order max pooling. When $\alpha<1$, the normalization function shrinks eigenvalues larger 1.0 and stretches  those less than 1.0. As  $\alpha$ (less than 1.0) decreases, the error rate continuously gets smaller until  the smallest value at around $\alpha=\frac{1}{2}$. With further decline of $\alpha$, however,  we observe the error rate grows consistently and soon is  larger than that of the plain cov pooling. Note that over the interval $[0.4,0.9]$ the performance of cov pooling varies insignificantly.  When $\alpha>1$, the effect of normalization is contrary, i.e., eigenvalues less than 1.0 are shrunk while those larger than 1.0 are stretched, which  is  not beneficial for covariance representations as indicated by the consistent growth of the  error rates. In all the following experiments, we set $\alpha=\frac{1}{2}$.

\vspace{3pt}\noindent\textbf{Impact of Various Normalization Methods.}\;\; We mainly compare three kinds of normalizations (i.e., MPN, M-Fro and M-$\ell_{2}$) as well as their combinations. Table~\ref{table:different-norm} summarizes the comparison results. Compared to  first-order pooling, the plain cov pooling  and cov pooling with normalizations decrease the errors by a gap of over $1.1\%$, which clearly indicates that the second-order pooling is better than the most commonly used, first-order pooling. When used separately,  all of the three matrix normalizations improve over the plain cov pooling.  MPN outperforms M-Fro and M-$\ell_{2}$ by ${\sim}1.3\%$ and ${\sim}1.1\%$, respectively,  but combinations of M-Fro (or MPN-$\ell_{2}$) with MPN  degrade the performance of separate MPN. We also made experiments by performing element-wise power normalization right after MPN , which, however, degraded the performance by over $1\%$. As our MPN-COV explores the second-order statistics, which is more complex than the first-order one  and may complicate minimization of the loss function, we consider warm initialization. Specifically, the weights of all layers before the cov pooling  are initialized with the trained network with max pooling and their learning rates are set to $\mathrm{logspace}[-2,-5, 20]$, while those after cov pooling are randomly initialized with learning rates doubled.  Compared to random initialization (training from scratch),  the warm initialization  decreases error by ${\sim}1.1\%$, which indicates it  helps MPN-COV network converge to a better local minimum of the loss function. In the end,  we compare in Table~\ref{table:different-norm-comparision-related} our MPN-COV network with two ConvNets with similar architectures, i.e.,  VGG-F~\cite{DBLP:conf/bmvc/ChatfieldSVZ14} and Krizhevsky et al.~\cite{nips2012cnn}. The result of VGG-F is obtained by us using  \href{http://www.vlfeat.org/matconvnet/models/imagenet-matconvnet-vgg-f.mat}{the model} released  on the MatConvNet website. Our MPN-COV network performs much better than both of them. 

\begin{table}[tb]
\renewcommand{\baselinestretch}{1.1}
\small
\centering
\begin{minipage}[t]{1.0\linewidth}
\centering
\begin{tabular}{llll}
\hline
method & init. & top-1 error   &  top-5 error \\
\hline
Plain COV (ours)                           & random   &  40.41  & 18.94\\ 
\hline
MPN-COV  (ours)                           & random   &  \textbf{38.51} & \textbf{17.60}\\
B-CNN~\cite{lin2015bilinear}      & random  &  39.89 & 18.32 \\
DeepO$_{2}$P~\cite{Ionescu_2015_ICCV} & random &  42.16 & 19.62  \\
\hline
\end{tabular}
\end{minipage}
\renewcommand{\baselinestretch}{1.0}
\caption{Error~(\%, \textit{1-crop}) comparison of MPN-COV  with two existing second-order pooling methods under AlexNet architecture.}
\label{table:comparison-with-existing}
\end{table}

\vspace{3pt}\noindent\textbf{Comparison with Existing Second-order Methods.}\;\; Here we compare with DeepO$_{2}$P~\cite{Ionescu_2015_ICCV} and B-CNN~\cite{lin2015bilinear}, two existing end-to-end, second-order pooling methods, neither of which has been previously  evaluated  on ImageNet dataset.
For  DeepO$_{2}$P, Ionescu et al. suggested implementation of the nonlinear, matrix logarithm  using  SVD and double-precision floating-point format for both the forward and backward propagations.  We adopted  \href{http://www.maths.lth.se/matematiklth/personal/sminchis/code/matrix-backprop.html}{the code} released by them. As suggested, we add $\epsilon=10^{-3}$ to the eigenvalues for numerical stability. For a  matrix $\mathbf{P}=[p_{ij}]$, B-CNN computes
$\mathbf{Q}=\big[\mathrm{sign}(p_{ij})(|p_{ij}|+\epsilon)^{\beta}\big]$ where  $\mathrm{sign}$ is the signum function.  For implementation, we adopted the~\href{http://vis-www.cs.umass.edu/bcnn/}{code} released by the authors of B-CNN, where  $\beta=0.5$ and $\epsilon=10^{-5}$ as suggested.     Table~\ref{table:comparison-with-existing} presents the comparison results. We can see that DeepO$_{2}$P is inferior to the plain cov pooling (no normalization). As analyzed in~Sec.~\ref{subsection:computational-perspective}, we attribute this to the fact  that logarithm is not suitable for the convolutional features given the distribution as shown in Fig.~\ref{subfigure:hist-and-norm-functions} (left), as it changes the order of eigenvalue significances. B-CNN slightly improves the plain cov pooling by ${\sim}0.5\%$, but outperformed by MPN-COV by $1.4\%$. We mention that we tuned $\epsilon$ for DeepO$_{2}$P and $\beta$ and $\epsilon$ for B-CNN but achieved trivial improvement.

\subsection{MPN-COV Under VGG-Net Architectures}\label{subsection-MPN-COV-VGG}

In this section, we combine MPN-COV   with two  VGG networks, i.e. VGG-M~\cite{DBLP:conf/bmvc/ChatfieldSVZ14} and VGG-16~\cite{Simonyan15}. We slightly modify VGG-M by presenting two configurations (config.). In config. a, we add  an additional $1\times 1\times 256$ conv. layer (filter size: $1\times 1$, channel: 256) right after Conv5, and for config. b, the numbers of channels of Conv3 and Conv4 are both raised from 512 to 640 while that of Conv5 is reduced to 256. Then our MPN-COV layer follows.  Note that the $1\times 1$  convolution   has been  used for  dimensionality  reduction and introducing nonlinearity~\cite{iclr2014_NIN,Szegedy_2015_CVPR}. Either config. produces a sample of features $\mathbf{X}\in \mathds{R}^{256\times 169}$ for the MPN-COV layer.  As seen from Table~\ref{table:MPN-COV-VGG-M}, under config. a with random initialization MPN-COV outperforms the first-order, max pooling  by ${\sim}2.5\%$ and  with warm initialization the gap increases to ${\sim}3.6\%$. For config. b the gains over max pooling are a little less than Config. a, i.e., ${\sim}2\%$  and ${\sim}3\%$ under random  and warm initialization, respectively. For VGG-16, we add a $1\times 1\times 256$ convolution after the last conv. layer, obtain the feature matrix $\mathbf{X}\in \mathds{R}^{256\times 196}$ for cov pooling. The results are shown in Table~\ref{table:MPN-COV-VGG-16}, from which we see that MPN-COV can bring  large improvement for VGG-16 architecture.
\begin{table}[tb]
\renewcommand{\baselinestretch}{1.1}
\small
\centering
\begin{minipage}[t]{1.0\linewidth}
\centering
\begin{tabular}{llll}
\hline
method & init. & config. a   &  config. b \\
\hline
First-order (ours) & Random & 37.07 & 37.31  \\
\hline
\multirow{2}{*}{MPN-COV (ours)}
& Random & 34.60    & 35.27      \\
& warm    & \textbf{33.44}   & 34.25            \\
\hline
\end{tabular}
\end{minipage}
\renewcommand{\baselinestretch}{1.0}
\caption{Top-1 error (\%, \textit{1-crop}) under VGG-M architecture. }
\label{table:MPN-COV-VGG-M}

\vspace{3pt}
\renewcommand{\baselinestretch}{1.1}
\small
\centering
\begin{minipage}[t]{1.0\linewidth}
\centering
\begin{tabular}{llll}
\hline
method    & init.    & top-1 error    & top-5 error  \\
\hline
First-order (ours) & random  &  29.62         & 10.81 \\
MPN-COV (ours)   & random & \textbf{26.55} & \textbf{8.94}       \\
\hline
\end{tabular}
\end{minipage}
\renewcommand{\baselinestretch}{1.0}
\caption{Error rates (\%, \textit{1-crop}) under VGG-16 architecture.}
\label{table:MPN-COV-VGG-16}

\vspace{10pt}
\renewcommand{\baselinestretch}{1.1}
\small
\centering
\begin{minipage}[t]{1.0\linewidth}
\centering
\begin{tabular}{lll}
\hline
method & top-1 error   &  top-5 error \\
\hline
MPN-COV (VGG-M) (ours)                            & \textbf{30.39} & \textbf{11.43}  \\
VGG-M~\cite{DBLP:conf/bmvc/ChatfieldSVZ14}  & 34.00 & 13.49       \\
\hline
Zeiler \& Fergus~\cite{Zeiler2014}          & 37.5  & 16.0  \\ 
OverFeat~\cite{iclr2014_OverFeat}           & 35.60  & 14.71  \\ 
\hline
\end{tabular}
\end{minipage}
\renewcommand{\baselinestretch}{1.0}
\caption{Error~(\%, \textit{10-crop}) comparison of MPN-COV (VGG-M)  with  ConvNets sharing similar architecture.}
\label{table:comparision-similar-with-VGG-M}

\vspace{6pt}
\setlength\tabcolsep{5pt}
\renewcommand{\baselinestretch}{1.1}
\small
\centering
\begin{minipage}[t]{1.0\linewidth}
\centering
\begin{tabular}{llll}
\hline
method & \#layers & top-1 err.  &  top-5 err. \\
\hline
MPN-COV(VGG-16)(ours)            & 17       & \textbf{24.68} & \textbf{7.75}  \\
VGG-16~\cite{Simonyan15}    & 16       & 27.41 & 9.20    \\
\hline
GoogleNet~\cite{Szegedy_2015_CVPR}& 22 & -- & 9.15  \\ 
PReLU-net B~\cite{He_2015_ICCV}   & 22 & 25.53 & 8.13  \\ 
\hline
\end{tabular}
\end{minipage}
\renewcommand{\baselinestretch}{1.0}
\caption{Error~(\%, \textit{10-crop}) comparison of MPN-COV (VGG-16) with two ConvNets having comparable number of conv. layers.}
\label{table:comparision-similar-with-VGG-VD}
\end{table}

Table~\ref{table:comparision-similar-with-VGG-M} presents comparison of  MPN-COV (VGG-M) with the original VGG-M~\cite{DBLP:conf/bmvc/ChatfieldSVZ14},  Zeiler \& Fergus~\cite{Zeiler2014} and OverFeat~\cite{iclr2014_OverFeat}, all sharing similar network architecture. Our MPN-COV (VGG-M) shows much better performance than them. In  Table~\ref{table:comparision-similar-with-VGG-VD}, we can see that in terms of top-1 error, MPN-COV (VGG-16) outperforms the original VGG-16~\cite{Simonyan15} by ${\sim}2.7\%$, and in terms of top-5 error, it performs better than  GoogleNet~\cite{Szegedy_2015_CVPR}  and PReLU-net B~\cite{He_2015_ICCV}   by ${\sim}1.4\%$  and ${\sim}0.4\%$, respectively. PReLU-net C is similar to PReLU-net B but significantly increases \#channels of every filter~\cite{He_2015_ICCV}, producing slightly better performance than ours.
Note that as the authors did not report the 10-crop results of the original VGG-M and VGG-16, we obtain them by using the best-performing models, i.e., \href{http://www.vlfeat.org/matconvnet/models/imagenet-matconvnet-vgg-m.mat}{matconvnet-vgg-m.mat} and \href{http://www.vlfeat.org/matconvnet/models/imagenet-vgg-verydeep-16.mat}{vgg-verydeep-16.mat}, respectively.

\subsection{MPN-COV Under ResNet Architecture}\label{subsection:MPN-COV-ResNet}

Finally, we integrate  MPN-COV into ResNet-50 (baseline). To retain as many number of features as possible, we do not perform downsampling in conv5\_1, as done in the original network, for the last set of  building blocks (i.e. conv5\_x). Then we connect the last addition layer (with ReLU) to a $1\times 1\times 256$ conv. layer, followed by the MPN-COV  layer. As such, we have  a sample of features $\mathbf{X}\in \mathds{R}^{256\times196}$ for covariance estimation. Regarding the time (ms) taken per image, MPN-COV network vs baseline are 18.21 vs 14.37 for training and 5.8 vs 3.5 for inference, respectively. We observe MPN-COV network converges faster: training/validation error rates (\%) of MPN-COV vs baseline reach 37.0/34.3 vs 50.35/44.55 at epoch 30 and 18.02/23.19 vs 25.98/25.76 at epoch 60.   Table~\ref{table:MPN-COV-ResNet-50} shows that  MPN-COV produces ${\sim}2.2\%$ top-1 error (1-crop) less than the first-order average pooling. Table~\ref{table:comparision-similar-with-ResNet-50} shows that, compared to the original ResNets,  with 10-crop prediction our  MPN-COV network performs $1.65\%$ better than   ResNet-50, while outperforming ResNet-101 and being comparable to ResNet-152. By exploiting  second-order statistics we  achieve performance matching extremely deep ConvNets with much shallower one.

\begin{table}[tb]
\renewcommand{\baselinestretch}{1.1}
\small
\centering
\begin{minipage}[t]{1.0\linewidth}
\centering
\begin{tabular}{llll}
\hline
method     & init.   & top-1 error & top-5 error \\
\hline
First-order (ours) & random &         24.95         &  7.52\\
MPN-COV (ours)    & random & \textbf{22.73} & \textbf{6.54}        \\
\hline
\end{tabular}
\end{minipage}
\renewcommand{\baselinestretch}{1.0}
\caption{Error rates (\%,\textit{1-crop}) under ResNet-50 architecture.}
\label{table:MPN-COV-ResNet-50}

\vspace{3pt}
\renewcommand{\baselinestretch}{1.1}
\small
\centering
\begin{minipage}[t]{1.0\linewidth}
\centering
\begin{tabular}{llll}
\hline
method                          & top-1 error    &  top-5 error      \\
\hline
MPN-COV (ResNet-50)  (ours)           & \textbf{21.20} &         5.74                \\
ResNet-50~\cite{He_2016_CVPR}   &         22.85  &         6.71    \\
\hline 
ResNet-101~\cite{He_2016_CVPR}  &         21.75  &         6.05    \\ 
ResNet-152~\cite{He_2016_CVPR}  &         21.43  & \textbf{5.71}    \\ 
\hline
\end{tabular}
\end{minipage}
\renewcommand{\baselinestretch}{1.0}
\caption{Error~(\%, \textit{10-crop}) comparison of MPN-COV (ResNet-50) with the original ResNets.}
\label{table:comparision-similar-with-ResNet-50}
\end{table}

\section{Conclusion}

This paper proposed a matrix normalized covariance (MPN-COV)  method for exploring  the second-order statistics in large-scale classification. MPN-COV amounts to robust covariance estimation given a small number of large-dimensional features. It also approximately  exploits  the  geometry of the space of covariance matrices, while circumventing  the downside of the well-known Log-Euclidean metric.  Extensive experiments on ImageNet 2012 dataset showed that our MPN-COV networks achieved competitive gains over its counterparts using only  first-order information.  In future we will combine MPN-COV with the Inception architecture~\cite{Szegedy_2015_CVPR}, and study applications of MPN-COV to visual tasks such as object detection, scene categorization and fine-grained visual recognition.

{\small
\bibliographystyle{ieee}
\bibliography{egbib}
}

\end{document}